# Dataset and Benchmark for Urdu Natural Scenes Text Detection, Recognition and Visual Question Answering

Hiba Maryam*, Ling Fu*, Jiajun Song, Tajrian ABM Shafayet, Qidi Luo, Xiang Bai**, and Yuliang Liu

Huazhong University of Science and Technology, Wuhan, China
{hiba, lingfu, u202014943, shafayet, qdiluo, xbai, ylliu}@hust.edu.cn

**Abstract.** The development of Urdu scene text detection, recognition, and Visual Question Answering (VQA) technologies is crucial for advancing accessibility, information retrieval, and linguistic diversity in digital content, facilitating better understanding and interaction with Urdu-language visual data. This initiative seeks to bridge the gap between textual and visual comprehension. We propose a new multi-task Urdu scene text dataset comprising over 1000 natural scene images, which can be used for text detection, recognition, and VQA tasks. We provide fine-grained annotations for text instances, addressing the limitations of previous datasets for facing arbitrary-shaped texts. By incorporating additional annotation points, this dataset facilitates the development and assessment of methods that can handle diverse text layouts, intricate shapes, and non-standard orientations commonly encountered in real-world scenarios. Besides, the VQA annotations make it the first benchmark for the Urdu Text VQA method, which can prompt the development of Urdu scene text understanding.
The proposed dataset is available at:
https://github.com/Hiba-MeiRuan/Urdu-VQA-Dataset-/tree/main

## 1 Introduction

Text in natural scenes frequently contains valuable information that's helpful in the recovery of visual information and the understanding of image contexts. It is proven [21] that people tend to concentrate more on the text when they are shown an image with other items in addition to the text, highlighting the importance of text recognition in image interpretation. The growth of smartphone and portable device applications that involve identifying and understanding the text within natural scenes has significantly fueled interest and research within the fields of computer vision and pattern recognition. This interest spans a variety of applications, including image-based content analysis, support for individuals with visual impairments, guidance for autonomous robots, and real-time language translation, among others. The Visual Question Answering (VQA) task also holds paramount importance in comprehending both visual content and textual queries, reflecting a holistic understanding of image-based information. VQA has significant implications for fields such as image captioning, human-computer interaction, and accessibility technologies.

---

* Equal Contribution
** Corresponding Author

The Latin text has been the primary focus of the majority of approaches proposed for natural scene text detection, recognition, and visual question answering. Because of their intrinsic complexity, cursive scripts like Arabic and Urdu have not received as much attention as they should. As a result, precise methods for identifying cursive text in natural settings are still being developed. However, despite the progress in text detection and recognition, challenges persist, especially in Urdu. The difficulty arises from factors such as variations in text fonts, styles, colors, orientations, sizes, blur, complicated backgrounds, uneven lighting, and lighting conditions. Additionally, images of natural scenes may include elements like brick shapes, fences, buildings, and leaves, which can resemble text structures, complicating identification and detection.

One limitation for academics tackling the problems of natural scene text detection, recognition, and VQA is the lack of a standardized dataset for cursive text. The availability of robust and diverse datasets is crucial for training accurate and effective recognition models. Additionally, a notable gap exists as there is currently no dedicated Urdu text VQA dataset, impeding the development and evaluation of models specifically tailored for understanding and answering questions related to Urdu text within images. Consequently, the goal of this research project is to gather natural scene photos with text in Urdu, compile them into a benchmark multi-task dataset not limited to text detection and recognition but also VQA, and make it available to other researchers to enable advancements in the understanding and processing of Urdu text within the context of real-world scenes.

In this research endeavor, we propose a dataset featuring multi-task annotations, strategically designed to advance the development of Urdu Optical Character Recognition (OCR) technologies. The dataset encompasses a comprehensive set of annotations, addressing various tasks essential for the holistic understanding of Urdu text in diverse scenarios. This multi-task dataset incorporates additional annotation points, strategically designed to facilitate the development and evaluation of methods capable of addressing diverse challenges encountered in real-world scenarios. These challenges include variations in text layouts, intricate shapes, and non-standard orientations. The inclusion of VQA tasks further enriches the dataset, fostering the creation of OCR technologies that can not only recognize Urdu text but also comprehend and respond to queries about the content, context, and relationships within the visual scenes. Therefore, this multi-faceted dataset serves as a valuable resource for researchers and practitioners working on the advancement of Urdu OCR technologies, enabling them to develop more sophisticated and versatile solutions that align with the complexities of real-world Urdu scenarios.

## 2    Related Work

### 2.1    Scene Text Detection

A dataset for character extraction and detection from scene images containing Urdu text has been proposed recently. There are 600 Urdu text-scene photos in the dataset. For training and testing purposes, the text region is manually cropped from these images to create a dataset of 18,000 Urdu characters with 48 × 48 dimensions [6]. An Urdu-text dataset has recently been published for Urdu text detection and recognition in natural scene images. This dataset contains a huge variety of images in terms of text size, writing styles, aspect ratios, background complexities, and handwritten text [8]. In a recent study [13], a Differentiable Binarization (DB) module was offered that could carry out the segmentation network binarization procedure. When combined with an optimized DB module, a segmentation network may establish the thresholds for binarization adaptively, which improves text detection performance and streamlines post-processing on a simple segmentation network. In another study [14], they propose the Pixel Aggregation Network (PAN), which can effectively combine text pixels by predicting similarity vectors and implementing learnable post-processing. In [15] the researchers build FCENet using a backbone, feature pyramid networks (FPN), and basic post-processing using non-maximum suppression (NMS) and inverse Fourier transformation (IFT). The experimental findings demonstrate the superiority of the FCENet particularly on the difficult, highly curved text subset.

### 2.2    Scene Text Recognition

Several OCR-based methods have been developed in the past, but their effectiveness is predominantly observed in recognizing characters within scanned materials. These methods often fall short when applied to natural-scene text recognition. In a color quantization-based approach outlined in [2], pixel values are classified into text and background using k-means clustering. This method is initiated by analyzing the colors of both the text and the backdrop. Another method, detailed in [3], utilizes k-means clustering based on color information to differentiate text from background.

Despite these efforts, the accuracy of text recognition in natural scenes remains a challenge. Recent advancements in binarization techniques and pre-processing procedures offer potential improvements, yet identification accuracy still lags behind expectations. Factors such as inconsistent illumination, complex backgrounds, and variations in text size, font, color, and orientation continue to pose challenges in achieving optimal results for text recognition in natural settings. Aside from English, much effort has been expended on Chinese text extraction. Chinese text extraction, for example, has been achieved using MSER-based algorithms in conjunction with clustering methodologies [4]. While there has been no substantial work on extracting Urdu text from natural scene photographs, it is nevertheless important to summarize the work done for several comparable languages, such as Arabic, Persian, and Uyghur.

Another study [5] described a text recognition method for Arabic text in scene photos that uses convolutional neural networks. After a total of 2,700 characters were retrieved by segmentation for 27 distinct characters (i.e., 100 photos per character), the work primarily addressed the recognition of isolated characters of Arabic text inside scene photographs. The authors in [6] have reported an attempt to recognize handwritten characters in Urdu. Nevertheless, text detection in photos of natural scenes is not covered by this work; instead, it focuses on the recognition of individual characters against a clear background.

Various classifiers are trained using HOG features, and their performance on the suggested dataset is assessed. The authors [7] report on yet another significant piece of work on Urdu text recognition. According to the background studies, a minimum number of research works are performed for detecting tiny objects using deep learning models or advanced image processing approaches. ASTER is presented as an end-to-end neural network model that consists of a recognition network and a rectification network in a different study [16]. Another study [17] presents a Transformer-inspired innovative architecture called the Self-Attention Text Recognition Network (SATRN) for text recognition of arbitrary shapes. To represent the two-dimensional (2D) spatial interdependence of characters in a scene text image, SATRN makes use of the self-attention mechanism. This work [18] proposes an autonomous, bidirectional, and iterative ABINet for scene text recognition that delivers state-of-the-art performance on various popular benchmarks and outperforms other models on low-quality photos.

### 2.3   Visual Question Answering (VQA)

The goal of Visual Question Answering (VQA) is to respond to a given natural language query regarding the picture. The recent studies on question answering for bar charts and diagrams [9][10], the work on machine-printed document images [11], and the work on textbook question answering [12] are all relevant to the task suggested in this paper. With a backdrop of text, graphs, and images, the Textbook Question Answering (TQA) dataset seeks to deliver multimodal answers; nevertheless, the textual data is presented in a machine-readable style. This isn't the case for the datasets' diagrams and charts presented, indicating that models need text recognition of some kind to complete these kinds of quality assurance tasks.

Although VQA tasks are becoming increasingly important, it is notable that there is currently a lack of a specialized Urdu text-VQA dataset. Taking note of this deficit, our suggested dataset takes on paramount significance as it attempts to bridge this gap by offering a benchmark and comprehensive Urdu text-VQA dataset.

## 3  Proposed Multi-task Urdu Dataset

Urdu is written in a modified version of the Arabic script, known as the Persian-Arabic script, which has been adapted to represent the unique phonological features of the language. This script is written from right to left, and its alphabet includes a combination of letters from Arabic and Persian, as well as additional characters specific to Urdu. Urdu script is a form of calligraphic writing where the letters are joined together to create words, and it is traditionally inscribed from right to left. The Urdu alphabet consists of 37 letters. These letters include 35 basic consonants and two additional letters that represent specific sounds in Urdu. Unlike in English, where each letter is written separately, many letters in Urdu are joined together when forming words. Some letters in Urdu have different shapes based on their position within a word. Diacritics are small marks placed above or below letters to indicate vowel sounds. In Urdu, short vowel sounds are usually not represented in writing, but long vowels and certain vowel combinations are indicated using diacritics.

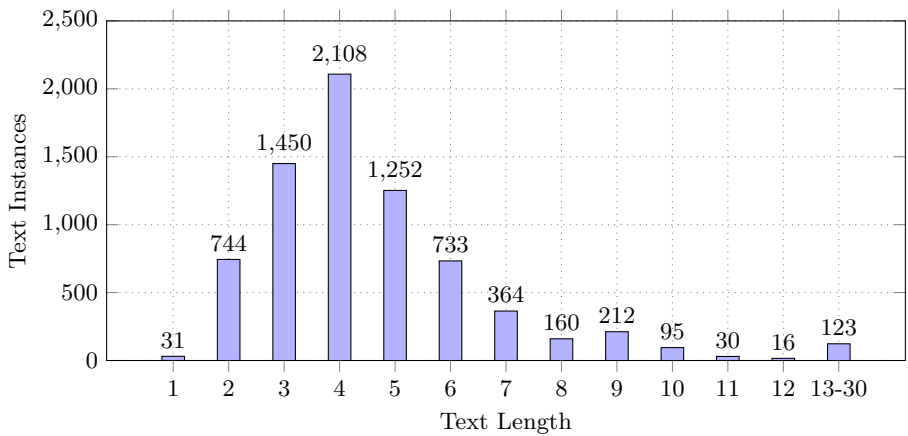

Fig. 1: Text Instances Counts by Text Length

### 3.1  Dataset Description

We have obtained high-resolution photographs of Urdu text scenes with resolutions higher than $1080 \times 800$, including street names, business names, advertisement banners, and road signboards. This will enable us to identify and detect Urdu text in natural-scene images. Because letters in cursive text, like Urdu, vary in location within the word and are connected, it is particularly challenging to break a word into separate characters.

These images were captured using a smartphone camera that lacks pre-processing filters, ensuring the authenticity of the visual content. The photographs were taken at different times throughout the day, intentionally capturing a diverse range of lighting conditions. This deliberate variation in lighting allows for the creation of a dataset that authentically represents Urdu text in real-world settings. The inclusion of photos taken under different lighting conditions enhances the robustness and applicability of the dataset for research and development purposes. The statistical analysis for the number of images with different text lengths is shown in Fig.1.

We manually labeled each word from these pictures and created a collection of characters from natural scenes in Urdu, as seen in Fig. 2.

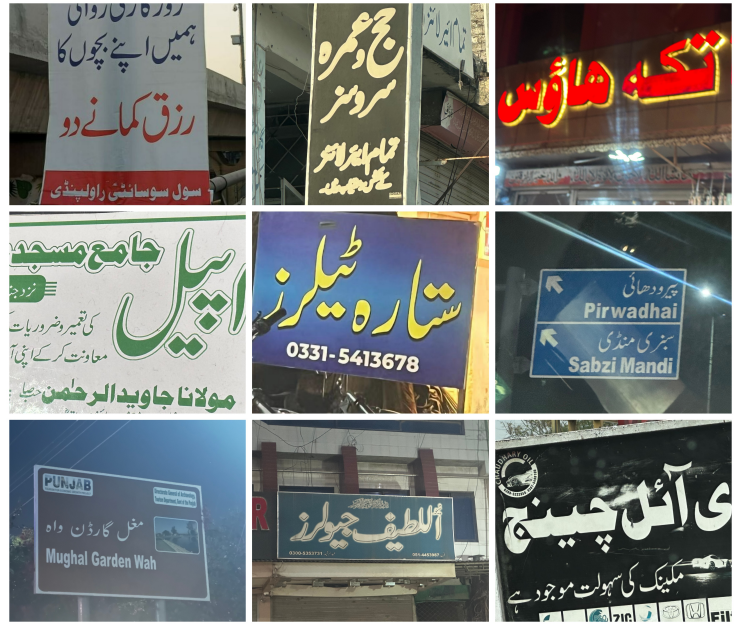

Fig. 2: Sample of images from the dataset

### 3.2   Text Detection and Recognition Annotation

Since proper annotation of text has a major impact on evaluation system performance, it is an essential component in the field of text detection. Many automatic or semi-automatic annotating tools have been created to obtain accurate annotations. These tools are intended to function well with Latin-script languages. However, these automatic technologies have difficulties delivering precise annotations for languages like Urdu or other cursive scripts. Cursive scripts sometimes have intricate and entwined characters. This is especially true when working with scripts like Urdu, which have complex shapes and varied layouts.

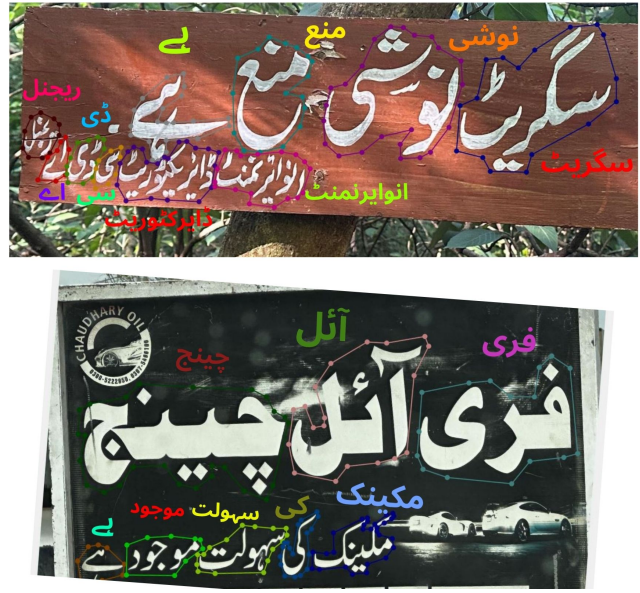

Fig. 3: Detection and Recognition Annotation details of Urdu-Text

Consequently, a manual approach is used to guarantee the accuracy and dependability of annotations. Every image in the collection is manually annotated, with a single enclosing polygon bounding box being used to annotate every text location. These annotated bounding boxes precisely capture all of the Urdu text instances. Even though human annotation takes more time, it becomes essential to preserve the annotated dataset's quality and accuracy for efficient text identification and recognition in languages like Urdu with complex scripts. In each image, the details of text instances are documented in a distinct .json file. This file includes lines where each one provides the bounding box details for an individual word along with the word itself. The data points represent the vertices of polygons, and the accompanying label denotes the word's textual information. The type of the text instance is shown in Fig. 3.

### 3.3  Visual Question Answering (VQA) Annotation

To construct our VQA dataset, we adopted a strategy that balanced speed with quality. The first step involved Gemini, a multimodal large language model capable of generating data from visual inputs and text prompts. Gemini's ability to create a substantial volume of question-answer pairs quickly provided a robust foundation for our dataset. However, while this automated approach was efficient, it required additional processing to ensure the generated data's accuracy and relevance, especially in the context of the Urdu language.

After the initial automated generation, we implemented a rigorous quality control process. This step was crucial because automated systems, while efficient, might not always capture the nuances of language and culture accurately. A team of native Urdu speakers meticulously reviewed the preliminary data produced by Gemini, focusing on linguistic accuracy, cultural context, and coherence. They corrected errors, adjusted awkward phrasing, and ensured that the questions and answers made sense in a natural Urdu context. This manual review process added a layer of human judgment that is often necessary for high-quality data annotation.

One of our key objectives was to make the dataset accessible to a broad research community. To achieve this, we carefully annotated every question in both Urdu and English. This dual-format approach supports cross-language analyses and comparisons, providing a versatile resource for various research purposes. It also allows researchers who might not be fluent in Urdu to work with the dataset, thus promoting broader accessibility and collaboration. The bilingual nature of the dataset is particularly valuable in the context of multilingual studies and helps ensure that our work has a wider impact.

The final dataset is a testament to the effectiveness of our two-step process. Combining the efficiency of Gemini's automated generation with the nuanced corrections from native Urdu speakers, the dataset is both extensive and reliable. It accommodates a wide range of queries and supports detailed analyses, whether researchers are interested in linguistic, cultural, or contextual aspects of VQA. The annotations in both Urdu and English add to the dataset's flexibility, making it a useful tool for a diverse range of applications. The results for the VQA annotations, as shown in Fig. 4, demonstrate the thoroughness of our approach and the high quality of the final product.

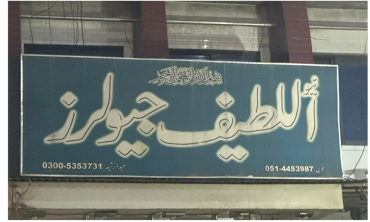
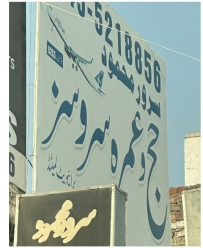

**question**: "What does the sign say?",
**answer**: Al-Lateef Jewellers (English),
**answer:** اللطیف جیولرز (Urdu)

**question**": "What is the Urdu text on the white signboard?",
**answer**":  Hajj and umrah services, (English)
**answer:** حج وعمرہ سروسز (Urdu)

Fig. 4: VQA Annotation details of Urdu-Text

## 4  Experiments

### 4.1  Implementation Details

For the experiments, there are 797 images for training and 248 images for testing from our proposed dataset. We evaluate our dataset with DBNet, PANet, and FCENet. Besides, we pre-train recognition methods with Cursive-Text, which contains 35,990 text images. We test the performance of some advanced text recognition methods, including ASTER, SATRN, and ABINet, on the proposed dataset. As for the VQA benchmark, we provide 3163 training question-answer pairs and 1043 testing pairs. The models were trained on one RTX 3090 GPU with a batch size of 16 for DBNet, PANet, and FCENet, 1024 for ASTER, 128 for SATRN, and 192 for ABINet. For optimization, SGD was used for DBNet and FCENet, Adam for PANet, ASTER, SATRN, and AdamW for ASTER. The initial learning rates were set to $7 \times 10^{-3}$ for DBNet with a Poly decay strategy, $1 \times 10^{-4}$ for PANet, $3 \times 10^{-3}$ for FCENet with decay by a factor of 0.8 every 200 epochs, $4 \times 10^{-4}$ for ASTER, $3 \times 10^{-6}$ for SATRN, and $1 \times 10^{-6}$ for ABINet.

In our experiment, we employed a two-pronged approach to tackle the tasks of text detection and text recognition. For the text detection task, we utilized the pre-training models provided by the MMOCR. These models were fine-tuned on our proprietary training set to yield the final results. Specifically, our implementation involves the utilization of DBNet with the pre-trained model `dbnet_-resnet18_fpnc_1200e_totaltext-3ed3233c.pth`, PANet with the pre-trained model `panet_resnet18_fpem-ffm_600e_ctw1500_20220826_144818-980f32d0_-.pth`, and FCENet with the pre-trained model `fcenet_resnet50_fpn_1500e_-totaltext91bd37af.pth`, each tailored to meet the unique demands of their respective tasks.

As for the text recognition task, we pretrain the text recognizers on Cursive-Text, and then we finetune and evaluate the recognizers on the proposed dataset.

### 4.2  Text Detection Experiment

The outcome presented here pertains to the findings derived from a text detection experiment conducted as part of this study. The experimental setup involved a dataset comprising a total of 797 images earmarked for training and 248 images allocated for testing. In the pursuit of evaluating the efficacy of text detection, three distinct methods were employed: DBNet, PANet, and FCENet. These are the distinct neural network designs used for text identification tasks in computer vision. Each of these systems leverages distinct procedures and strategies for properly detecting text occurrences within images. The Comparison between DBNet, PANet, and FCENet is shown in the table 1.

In the realm of text detection, the metric employed for assessing performance is denoted as hmean. This metric is utilized to provide a comprehensive evaluation that considers both precision and recall, contributing to a more nuanced understanding of the model's effectiveness in detecting text instances within the given dataset. The utilization of hmean ensures a robust evaluation framework, considering the interplay between precision and recall in the context of text detection methodologies. The experimental results for FCENet are shown in Fig. 5.

Table 1: Comparison of DBNet, PANet, and FCENet

| Model | Architecture | Advantages | Disadvantages |
|-------|--------------|------------|---------------|
| DBNet[13] | Differentiable binarization | High accuracy; robust against complex backgrounds | Requires additional post-processing |
| PANet[14] | Path Aggregation Network | Good for multi-scale text detection | Limited capability with highly curved text |
| FCENet [15] | Fourier Contour Embedding | Robust with arbitrary-shaped text | Computationally intensive |

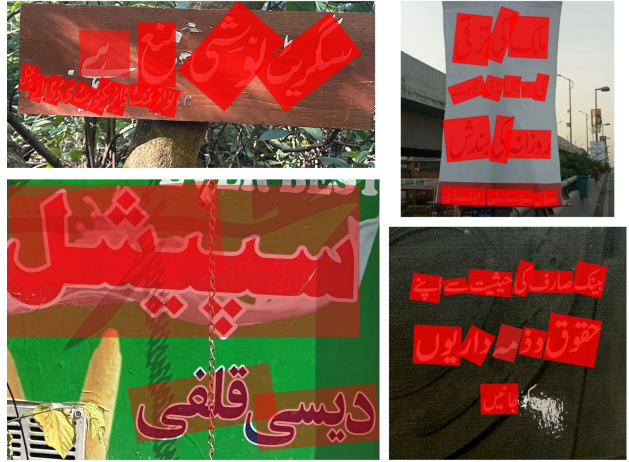

Fig. 5: Text Detection Results for FCENet

## 4.3   Text Recognition Experiment

The provided text recognition results for SATRN in Fig. 6 pertain to the experimental evaluation of text recognition. Table 2 shows the comparison of different recognition methods on the test set of our proposed dataset. These methods were evaluated on the test dataset to gauge their inherent capabilities for recognizing text.

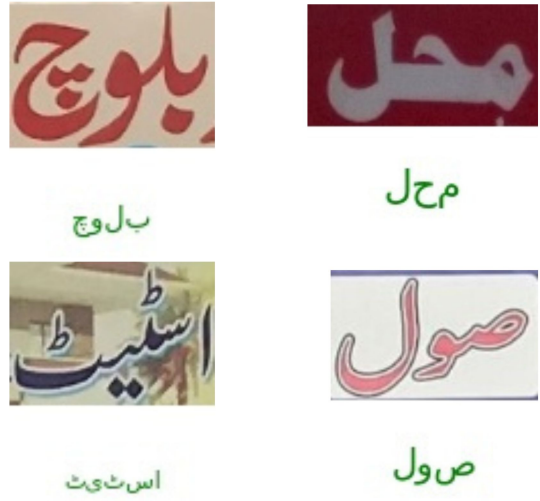

Fig. 6: Text Recognition Results for SATRN

Table 2: Comparison of different detection methods on the test set of our proposed dataset results on benchmarks. 'P', 'R', and 'F' are denoted as Precision, Recall, and F-measure, respectively. F-measure is the major evaluation metric.

| Methods | Venue | Backbone | Test Set | | |
|---------|-------|----------|------|------|------|
| | | | **P** | **R** | **F** |
| DBNet[13] | AAAI'2020 | Res18 | 62.6 | 61.2 | 61.9 |
| PANet[14] | CVPR'2018 | Res18 | 53.4 | 46.0 | 49.4 |
| FCENet[15] | CVPR'2021 | Res50 | 64.9 | 62.9 | 63.9 |

Table 3 offers a comparative overview of various recognition methods evaluated on the test set of the proposed dataset. These methods underwent evaluation on the test dataset to ascertain their inherent proficiency in recognizing text.

Table 3: Comparison of ASTER, SATRN, and ABINET

| Model | Architecture | Strengths | Weaknesses |
|---|---|---|---|
| **ASTER**[16] | Convolutional RNN with Attention | Robust for various text orientations and styles | Higher training complexity due to RNN layers |
| **SATRN**[17] | Transformer-based with multi-head self-attention | Strong for long text sequences | Requires extensive training data and resources |
| **ABINET**[18] | Transformer-based with Adaptive Bidirectional Encoding | Handles sequential and non-sequential text patterns | Complex training due to adaptive encoding layers |

Table 4: Comparison results of different recognition methods on the test set of our proposed dataset

| Methods | Venue | Test Set |
|---|---|---|
| ASTER[16] | TPAMI'2019 | 74.0 |
| SATRN[17] | CVPRW'2021 | 75.0 |
| ABINet[18] | CVPR'2021 | 67.2 |

## 4.4   Visual Question Answering (VQA) Experiments

In our VQA experiments, the lack of sufficient Urdu training data posed significant challenges. The multimodal large model is not trained on enough Urdu data, so the correlation performance is poor. This scarcity is a key reason for the relatively poor performance of Gemini when evaluated on our proposed dataset. Despite their versatility in handling multilingual tasks, these models often struggle when the training data in a specific language is limited or absent. This issue was evident in our experiments, where the generated answers frequently lacked accuracy and contextual relevance.

Due to the complexity of Urdu language recognition, multimodal large models often can only recognize a small portion of the content. If conventional evaluation methods are used, the evaluation scores of multimodal large models will be very low. Therefore, the threshold restrictions were removed in the experiment to observe the performance of the multimodal large model on this dataset

We used the Gemini model to evaluate our dataset due to its open-source nature and its capacity to process multiple languages, including Urdu. For our experiments, we presented Gemini with both textual questions and corresponding images, asking it to generate responses in English. This setup allowed us to assess Gemini's ability to comprehend visual content and its proficiency in generating linguistically accurate responses in English. We then used the Average Normalized Levenshtein Similarity (ANLS) [20] metric to measure the similarity between Gemini's generated answers and the ground truth, providing a quantitative measure of performance.

By evaluating Gemini on our VQA dataset, we aimed to better understand the model's limitations and identify areas for improvement. The results of our experiments demonstrated that while current large language models struggle with the complexity and nuances of Urdu-based VQA tasks, the dataset serves as a valuable resource for ongoing research and development in this area.

Table 5: VQA experiment result *Average Normalized Levenshtein Similarity (ANLS)

| Method | ANLS* |
|--------|-------|
| Gemini[19] | 46.3 |

## 5   Conclusion and Future Work

This paper introduces a new dataset comprising natural scene images with Urdu text annotations, specifically designed for text detection, recognition, and Visual Question Answering (VQA) tasks. The dataset addresses the limitations of existing datasets by providing diverse, high-resolution images with fine-grained annotations for text instances, including complex shapes, orientations, and backgrounds. Additionally, we propose a new task for Urdu scene text understanding, namely Urdu Text-VQA, which combines computer vision and natural language processing to answer questions about images containing Urdu text. Experimental results on the proposed dataset demonstrate the effectiveness of the dataset in evaluating Urdu scene text analysis methods and highlight the challenges in developing accurate and robust algorithms for text detection, recognition, and VQA in Urdu natural scene images.

As part of future work, we plan to expand the dataset by including more diverse images and annotations to further improve the performance and generalization of text detection, recognition, and VQA models.

Additionally, we aim to explore advanced techniques such as deep learning and reinforcement learning to develop more accurate and robust algorithms for Urdu scene text analysis. Moreover, we plan to investigate other related tasks such as text-based image retrieval and text-to-speech synthesis to enhance the usability and accessibility of Urdu scene text data for a wide range of applications.

**Acknowledgments** This work was supported by the National Natural Science Foundation of China (No.62225603, No.62206104).